\def\year{2019}\relax
\documentclass[letterpaper]{article} 
\usepackage{aaai19}  
\usepackage{times}  
\usepackage{helvet}  
\usepackage{courier}  
\usepackage{url}  
\usepackage{graphicx}  
\usepackage{booktabs}       
\usepackage{bm}
\usepackage{amssymb}
\usepackage{algorithm,algorithmic}
\usepackage{amsmath,multirow,colortbl}

\title{Improving Image Captioning with Conditional Generative Adversarial Nets}
\author{Chen Chen, Shuai Mu, Wanpeng Xiao, Zexiong Ye, Liesi Wu, Qi Ju\\
Tencent AI Lab, Shenzhen, China 518000 \\
\{beckhamchen, harrymou, wanpengxiao, joeyye, henrylwu, damonju\}@tencent.com\\
}

\begin{document}
%
\maketitle
\begin{abstract}
In this paper, we propose a novel conditional-generative-adversarial-nets-based image captioning framework as an extension of traditional reinforcement-learning (RL)-based encoder-decoder architecture. To deal with the inconsistent evaluation problem among different objective language metrics, we are motivated to design some ``discriminator'' networks to automatically and progressively determine whether generated caption is human described or machine generated. Two kinds of discriminator architectures (CNN and RNN-based structures) are introduced since each has its own advantages. The proposed algorithm is generic so that it can enhance any existing RL-based image captioning framework and we show that the conventional RL training method is just a special case of our approach. Empirically, we show consistent improvements over all language evaluation metrics for different state-of-the-art image captioning models. In addition, the well-trained discriminators can also be viewed as objective image captioning evaluators.
\end{abstract}

\section{Introduction}\label{sec:intro}

Generating natural language descriptions of given images, known as image captioning, has attracted great academic and industrial interest since it can be widely used in image-text cross-searching, early childhood education and eye-handicapped people assistance. Compared with other computer vision tasks, \textit{e.g.} image classification, object detection and semantic segmentation, image captioning is a more challenging and comprehensive task --- as it requires a fine-grain understanding of image objects as well as their attributes and relationships. Therefore, image captioning can be viewed as an interdisciplinary research domain of computer vision and natural language processing (NLP).

Inspired by the successful application of the encoder-decoder paradigm in neural machine translation (NTM) with RNNs, some pioneer works~\cite{m-RNN,show_tell} creatively proposed to replace an RNN with a CNN to encode image features. Since then, the CNN-RNN structure has become a standard configuration of image captioning algorithms. Most prior works used maximum likelihood estimation (MLE) for training. However, as pointed out in~\cite{sequence_level_training}, this approach suffers from error accumulation, namely a bias exposure problem~\cite{scheduled_sampling}, which creates a mismatch between training and testing --- since at test-time,
the model uses the previously generated words from the model distribution to predict the next word while directly uses ground-truth during training.

In order to address the exposure bias issue, many works incorporated reinforcement learning (RL) into the training stage to directly optimize language metrics such as BLEU~\cite{BLEU}, METEOR~\cite{METEOR}, ROUGE~\cite{ROUGE}, CIDEr~\cite{CIDEr} and SPICE~\cite{SPICE}. A policy-gradient~\cite{PolicyGradient}-based RL method was first employed in image captioning, such as REINFORCE with a baseline~\cite{image_caption_self-critical} and Monte-Carlo rollouts~\cite{ImageCaptionPolicyGradientSpider}. Then, the so-called actor-critic RL algorithm was applied in~\cite{ImageCaptionDeepReninforcement,ImageCaptionActorCritic}, which trains a second ``critic'' network to estimate the expected future reward during the intermediate state when generating each word given the policy of an actor network.

Recently, the most popular generative model--generative adversarial nets (GANs)~\cite{GANs}--has achieved great success in computer vision tasks. But unlike the deterministic continuous mapping from random vectors to the image domain, many NLP tasks are discrete domain generation issues. Inspired by the idea of reinforcement learning, a SeqGAN algorithm~\cite{SeqGAN} was proposed to bypass the generator differentiation problem by directly performing the gradient policy update and successfully applied the result to text and music generation. In~\cite{CondGANImageCaption} and~\cite{GumbelSamplerImageCaption}, the authors conducted the first study that explores the use of conditional GANs~\cite{CondGAN} or adversarial
training in combination with an approximate Gumbel sampler in generating image descriptions. However, what these papers concern about most is naturalness and diversity of descriptions while sacrificing the fidelity, which results in much lower language metrics scores compared with other image captioning algorithms.

In order to achieve high evaluation scores over objective automatic metrics, many prior works chose to directly optimize one metric (BLEU, CIDEr, SPICE, \textit{etc.}) or combination of them. However, optimizing one metric cannot ensure consistent improvements over all metrics. Therefore, in order to simultaneously improve all language evaluation metrics and generate more human-like descriptions, we are motivated to design a ``discriminator'' network to judge whether the input sentence is human described or machine generated based on the idea of conditional GANs~\cite{CondGAN}. In this paper, we propose to employ an adversarial training method to alternatively improve the generator (caption generation network) and the discriminator (sentence evaluation network).

The main contributions of our proposed image captioning framework are listed as follows:
\begin{itemize}
  \item Our proposed algorithm is generic so that it can enhance any existing RL-based image captioning framework and we show consistent improvements over all evaluation metrics by experiments.
  \item The well-trained discriminators can also be viewed as evaluators for image captioning learned by neural network instead of the traditional handcrafted evaluation metrics.
  \item Based on the up-down~\cite{image_caption_Top-Down} image captioning model, the improved results trained by our approach are uploaded to MSCOCO online test server and achieve the state-of-the-art performance.
\end{itemize}


\section{Image Captioning Via Reinforcement Learning}\label{sec:RL}

As described in the introduction, the traditional RNN model training method is MLE. That is, the model parameters $\bm{\theta}$ of the caption generator are trained to maximize
\begin{equation}\label{eq:MLE}
\begin{aligned}
  J_G(\bm{\theta}) &= \frac{1}{N}\sum_{j=1}^N\log G_{\bm{\theta}}(\bm{x}^j|\bm{I}^j) \\
                   &= \frac{1}{N}\sum_{j=1}^N\sum_{t=1}^{T_j}\log G_{\bm{\theta}}(\bm{x}_t^j|\bm{x}_{1:t-1}^j,\bm{I}^j),
\end{aligned}
\end{equation}
where $\bm{I}^j$ is the $j$-th image, $\bm{x}^j=$($\bm{x}_1^j,\bm{x}_2^j,\cdots,\bm{x}_{T_j}^j$) is the
ground truth caption of the $j$-th image, $T_j$ is the caption length of the $j$-th image, $N$ is the total number of training examples, and $G_{\bm{\theta}}(\cdot)$ is the probability of generated words given an image or previous words, parameterized by $\bm{\theta}$ (or we can directly call $G_{\bm{\theta}}$ the generator).

By using the RL terminologies as described in~\cite{RL}, in an encoder-decoder image captioning paradigm, the decoder can be viewed as an ``agent'' that interacts with
an external ``environment'' (input words and image features extracted by the encoder). The ``policy'' is the caption generator $G_{\bm{\theta}}$, that results
in an ``action'' that is the prediction of the next word. After taking each ``action'', the ``agent'' updates its internal ``state'' (weights of decoder, attention models, \textit{etc.}). Upon generating the end-of-sequence (EOS) token, the ``agent'' returns a ``reward'', denoted by $r$, that is, for instance, a language evaluation metric score (CIDEr, BLEU, SPICE, \textit{etc.}) calculated by comparing generated sentences and the corresponding ground truth. So, the goal of RL training is to maximize the final expected reward of the generator:
\begin{equation}\label{eq:RL_reward}
L_G(\bm{\theta}) = \mathbb{E}_{\bm{x}^s\sim G_{\bm{\theta}}}\left[r(\bm{x}^s)\right],
\end{equation}
where $\bm{x}^s=(\bm{x}^s_1,\bm{x}^s_2,\cdots,\bm{x}^s_T)$ is a sample sequence from generator $G_{\bm{\theta}}$. In practice, the expected value is approximated by a Monte-Carlo sample~\cite{RL}:
\begin{equation}\label{eq:RL_reward_sample}
L_G(\bm{\theta})\approx r(\bm{x}^s),\quad \bm{x}^s\sim G_{\bm{\theta}}.
\end{equation}
Typically, the gradient $\nabla_{\bm{\theta}}L_G(\bm{\theta})$ can be calculated by a policy gradient approach such as  REINFORCE~\cite{REINFORCE} algorithm with a baseline function $b$ to effectively reduce the variance of the gradient estimate:
\begin{equation}\label{eq:RL_reward_gradient_sample}
\nabla_{\bm{\theta}}L_G(\bm{\theta})\approx \sum_{t=1}^{T_s}(r(\bm{x}^s_{1:t})-b)\nabla_{\bm{\theta}}\log G_{\bm{\theta}}(\bm{x}^s_t|\bm{x}^s_{1:t-1}).
\end{equation}

The baseline $b$ can be an arbitrary function, as long as it does not depend on the ``action'' $\bm{x}^s$. A self-critical sequence training (SCST)~\cite{image_caption_self-critical} method employs the reward $r(\bm{x}^g)$ obtained by the current model under the greedy decoding algorithm used at test time as the baseline function. Then, the gradient function can be written as
\begin{equation}\label{eq:RL_reward_gradient_sample_SCST}
\nabla_{\bm{\theta}}L_G(\bm{\theta})= \sum_{t=1}^{T_s}(r(\bm{x}^s)-r(\bm{x}^g))\nabla_{\bm{\theta}}\log G_{\bm{\theta}}(\bm{x}^s_t|\bm{x}^s_{1:t-1}).
\end{equation}
Since the reward of the sample sequence $\bm{x}^s_{1:T_s}$ and the greedy-decoded sequence $\bm{x}^g_{1:T}$ are the same during each time step $t$, we omit the subscripts of $\bm{x}^s$ and $\bm{x}^g$ in Eq.~(\ref{eq:RL_reward_gradient_sample_SCST}).

\section{Proposed Conditional Generative Adversarial Training Method}\label{sec:algorithm}

\subsection{Overall Framework}\label{subsec:overall}
\begin{figure*}[t]
  \centering
  \centerline{\includegraphics[width=17cm]{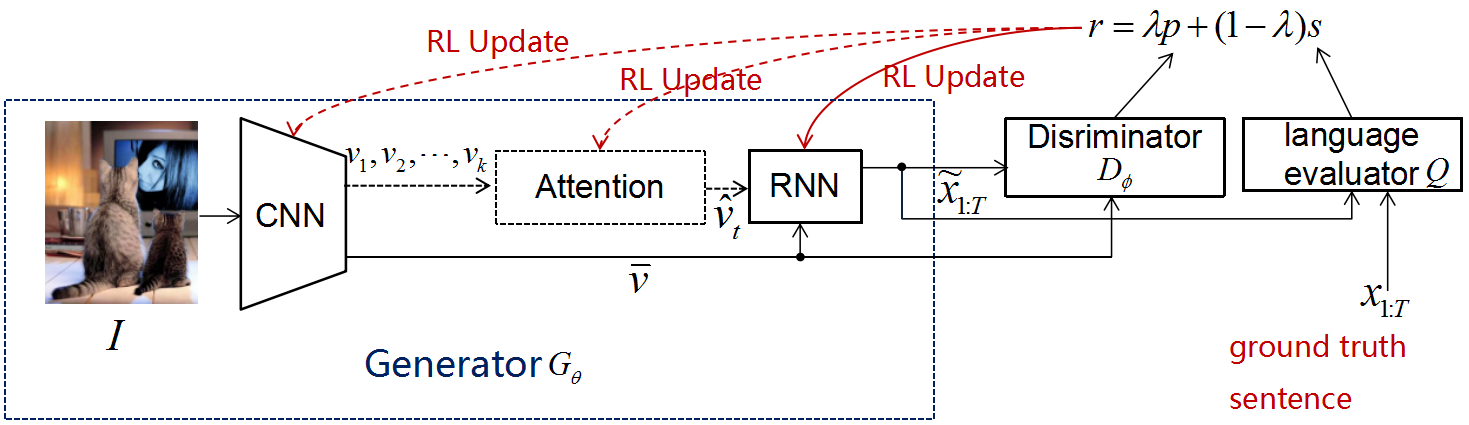}}
\caption{The overall framework of our proposed generative adversarial training method. The generator contains a CNN encoder, a RNN decoder and an unnecessary attention mechanism. The discriminator cooperated with the language evaluator provides feedback reward to update generator parameters by the reinforcement learning method (notice that CNN weights can be pre-trained and fixed).}
\label{fig:Overall_Framework}
\end{figure*}

As described in the introduction, the most commonly used image captioning model is the so-called encoder-decoder framework. Typically, a CNN is employed as the encoder and an RNN is utilized as the decoder. Together with the attention mechanism and reinforcement learning method, a general caption generation framework is shown in the left part of Fig.~\ref{fig:Overall_Framework}, denoted as the generator. Inspired by the well-known generative adversarial nets~\cite{GANs}, a discriminator can be embedded into the image captioning framework to further improve the performance of the generator, which is the initial design spirit of our proposed adversarial training method. Notice that our proposed framework is generic and can enhance any existing RL-based encoder-decoder model so the attention mechanism may be unnecessary. In addition, the CNN encoder can be pre-trained on other datasets so that its weights can be fixed during decoder training (see the dashed lines and squares in Fig.~\ref{fig:Overall_Framework}).

As shown in Fig.~\ref{fig:Overall_Framework}, after generating a sentence ${\tilde{\bm{x}}_{1:T}}$, two modules will compute two scores based on different criterions: a discriminator $D_{\bm{\phi}}$ with parameters $\bm{\phi}$ will produce a probability $p\in[0,1]$ that indicates a given sentence is human generated rather than machine generated, and a language evaluator module will calculate an objective score $s$ based on some predefined evaluation metrics $Q$ such as BLEU, CIDEr and SPICE. Notice that the discriminator will be improved together with the generator alternatively during training while the language evaluator module is a predefined function and is strictly fixed during training.
Therefore, the two modules are cooperated together to obtain a criterion that balances the fidelity (achieves a high score under objective evaluation metrics) and naturalness (a human-like language style).

Finally, the obtained reward for reinforcement learning after generating a full sentence $\tilde{\bm{x}}$ given image $\bm{I}$ and ground-truth sentence $\bm{x}$ is calculated as
\begin{equation}\label{eq:reward_function}
  r(\tilde{\bm{x}}|\bm{I},\bm{x})=  \lambda\cdot p +(1-\lambda)\cdot s = \lambda\cdot D_{\bm{\phi}}(\tilde{\bm{x}}|\bm{I}) +(1-\lambda)\cdot Q(\tilde{\bm{x}}|\bm{x}),
\end{equation}
where $\lambda$ is a hyper-parameter between 0 and 1.

In the following subsections, we will introduce two kinds of discriminators--CNN-based and RNN-based structures--in detail.

\subsection{CNN-based Discriminator Model}\label{sebsec:CNN_Discriminator}
\begin{figure*}[t]
\begin{minipage}[b]{.48\linewidth}
  \centering
  \centerline{\includegraphics[width=8.5cm]{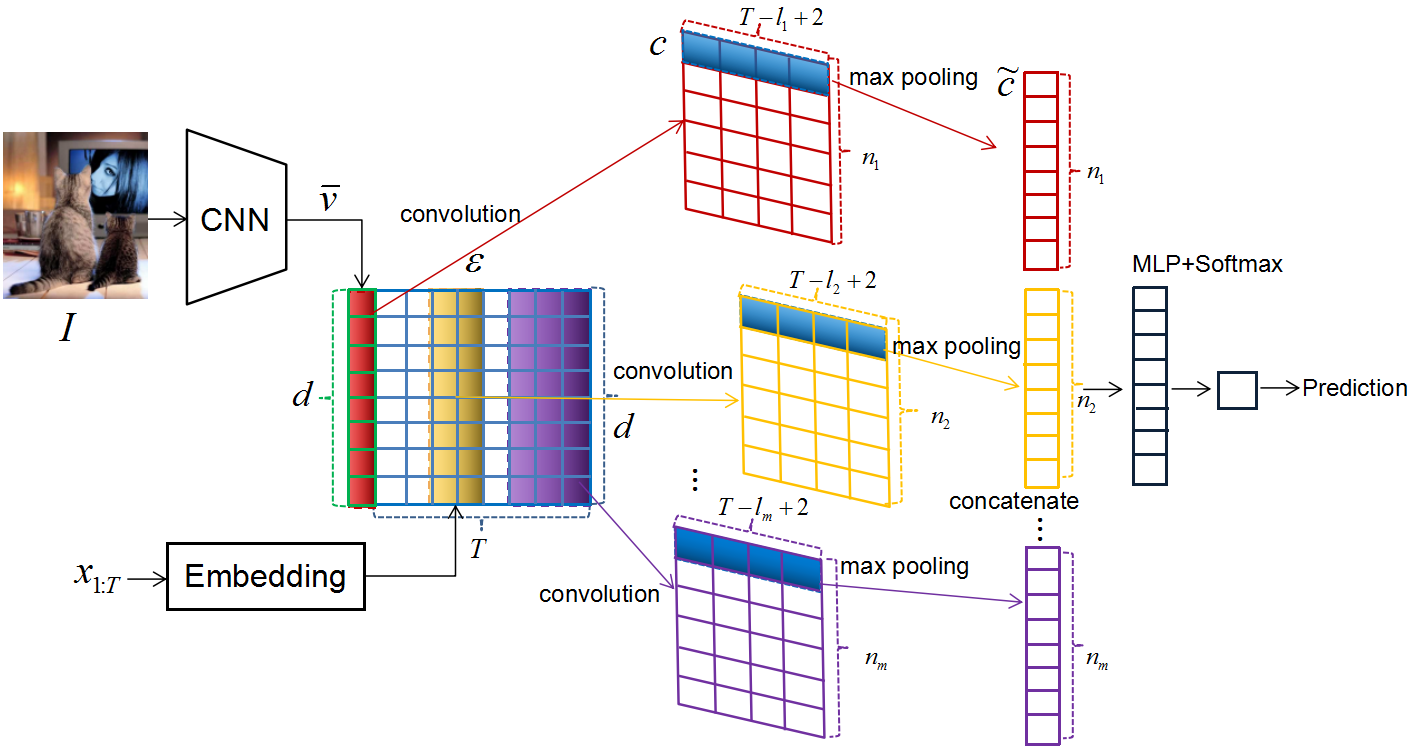}}
  \centerline{\small{(a) CNN-based discriminator}}
\end{minipage}
\hfill
\begin{minipage}[b]{0.48\linewidth}
  \centering
  \centerline{\includegraphics[width=8.5cm]{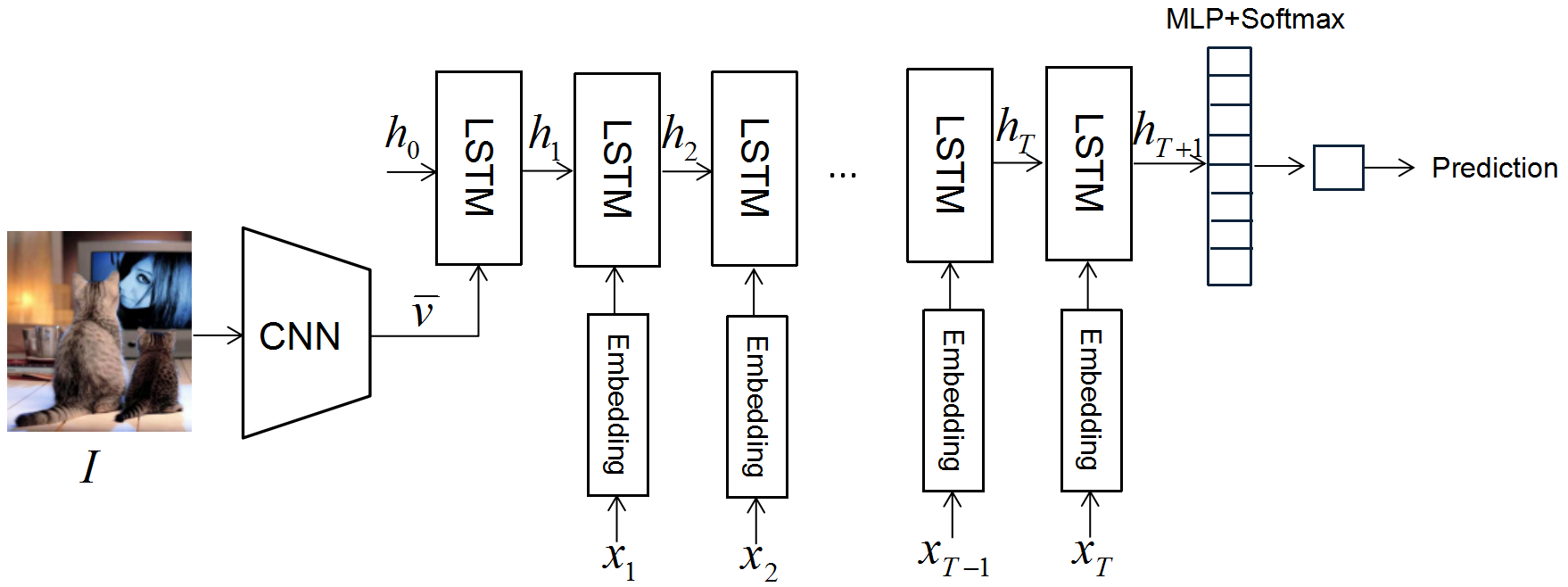}}
  \centerline{\small{(b) RNN-based discriminator}}
\end{minipage}
\caption{CNN and RNN-based discriminator architectures. Best viewed in colour.}
\label{fig:discrimiator}
\end{figure*}

\begin{table}[t]
\begin{center}
\caption{Kernel sizes and number of CNN discriminators} \label{tab:CNN_kernel_size_number}
\begin{tabular}{c}
\toprule[2pt]   (window size, kernel numbers)\\
\midrule[1pt] (1,100), (2,200), (3,200), (4,200), (5,200), (6,100),\\
(7,100), (8,100), (9,100), (10,100), (15,160), (16,160)\\
\bottomrule[2pt]
\end{tabular}
\end{center}
\end{table}

Recently, many CNN-based algorithms have shown great effectiveness in complicated NLP tasks such as sentence classification~\cite{CNNSentenceClass} and text classification~\cite{TextUnderstanding}. Therefore, in this subsection, we first present the conditional CNN as our discriminator for real or fake sentence classification.

Following the discriminator design of SeqGAN~\cite{SeqGAN}, as illustrated in Fig.~\ref{fig:discrimiator}(a), first, we should build a feature map that consists of image features and sentence features, such as
\begin{equation}\label{eq:image_sentence_feature}
\bm{\varepsilon} = \bm{\bar{v}}\oplus\bm{E}\cdot\bm{x}_1\oplus\bm{E}\cdot\bm{x}_2\oplus\cdots\oplus\bm{E}\cdot\bm{x}_T,
\end{equation}
where $\bm{\bar{v}}={\rm CNN}(\bm{I})$ is the $d$-dimensional image feature preprocessed by a CNN for input image $\bm{I}$, $\bm{E}\in \mathbb{R}^{d\times U}$ is the embedding matrix to map a $U$-dimensional one-hot word vector $\bm{x}_i$ $(i=\{1,2,\cdots,T\})$ into a $d$-dimensional token embedding, and $\oplus$ is the horizontal concatenation operation to build the matrix $\bm{\varepsilon}\in \mathbb{R}^{d\times(T+1)}$.

In order to extract different features, we apply $m$ group convolution kernels with different window sizes $d\times l_i$ $(i=\{1,2,\cdots,m\})$, each of which consists of $n_i$  $(i=\{1,2,\cdots,m\})$ kernels. The detailed design of kernel window sizes and number is presented in Table~\ref{tab:CNN_kernel_size_number} (slightly different from~\cite{SeqGAN}).  Without loss of generality, a kernel $\bm{w}\in \mathbb{R}^{d\times l}$ with window size $d\times l$
applied to the concatenated feature map of the input image and sentence will produce a feature map $\bm{c} = [c_1,c_2,\cdots,c_{T-l+2}]$.
Concretely, a specific feature is calculated as $c_i = {\rm ReLU}(\bm{w}\ast\bm{\varepsilon}_{i:i+l-1} + \bm{b})$, where $i=\{1,2,\cdots,T-l+2\}$, $\ast$ is the convolution operation, $b$ is a bias term, and ${\rm ReLU}(\cdot)$ is the Rectified Linear Unit.

Then we apply a max-over-time pooling operation over the feature map $\tilde c=\max\{\bm{c}\}$ and concatenate all the pooled features from
different kernels to form a feature vector $\bm{\tilde c}\in \mathbb{R}^n$ $(n=\sum_{i=1}^m n_i)$. Following the instruction of~\cite{SeqGAN},
to enhance the performance, we also add the highway architecture~\cite{HighwayNetworks} before the
final fully connected layer:
\begin{equation}\label{eq:high_way}
\begin{aligned}
  \bm{\tau} &= \sigma(\bm{W}_T\cdot\bm{\tilde c} + \bm{b}_T)\\
  \bm{H}    &= {\rm ReLU}(\bm{W}_H\cdot\bm{\tilde c} + \bm{b}_H)\\
  \mathbf{\tilde C} &= \bm{\tau}\odot\bm{H} + (\bm{1}-\bm{\tau})\odot\bm{\tilde c},
\end{aligned}
\end{equation}
where $\bm{W}_T,\bm{W}_H\in \mathbb{R}^{n\times n}$ and $\bm{b}_T,\bm{b}_H\in \mathbb{R}^{n}$ are high way layer weights and bias, respectively, $\sigma$ is the sigmoid function and $\odot$ is the piece-wise multiplication operation.

Finally, a fully connected layer and sigmoid transformation are applied to $\mathbf{\tilde C}$ to get the probability that a given sentence is real under a given image:
\begin{equation}\label{eq:CNN_prob_output}
p = \sigma(\bm{W}_o\cdot\mathbf{\tilde C}+\bm{b}_o),
\end{equation}
where $\bm{W}_o\in \mathbb{R}^{2\times n}$ and $\bm{b}_o\in \mathbb{R}^{2}$ are output layer weights and bias, respectively.

\subsection{RNN-based Discriminator Model}\label{subsec:RNN_Discriminator}
Since the most commonly used sequence modeling network is an RNN, in this subsection, we present the RNN-based discriminator architecture that consists of the standard LSTM structure, a fully connected linear layer and a softmax output layer.

For the first time step, the image feature vector $\bm{\bar{v}}$
is fed into the LSTM as an input with the randomly initialized hidden state $\bm{h}_0\in \mathbb{R}^{d}$. Then, for the following time steps, the input vectors will be changed to token embeddings $\bm{E}\cdot\bm{x}_t$ ($t=\{1,2,\cdots,T\}$).
The mathematical expressions are shown as follows:
\begin{equation}\label{eq:RNN_Discriminator}
\bm{h}_{t+1}=\left\{
                \begin{aligned}
                {\rm LSTM}(\bm{\bar{v}},\bm{h}_t) & & t=0 \\
                {\rm LSTM}(\bm{E}\cdot\bm{x}_t,\bm{h}_t) & & t=1,2,\cdots,T
                \end{aligned}
        \right..
\end{equation}
Then, after a fully connected layer and softmax layer, the probability that a given sentence is real under a given image can be calculated as:
\begin{equation}\label{eq:RNN_prob_output}
p=\sigma(\bm{W}_R\cdot\bm{h}_{T+1}+\bm{b}_R),
\end{equation}
where $\bm{W}_R\in \mathbb{R}^{2\times n}$ and $\bm{b}_R\in \mathbb{R}^{2}$ are linear layer weights and bias, respectively.
\subsection{Algorithm}
In order to incorporate the conditional GAN idea into the reinforcement learning method, the necessary back-propagated signal is the final reward $r$, as depicted in Fig.~\ref{fig:Overall_Framework}. Since the language evaluation score $s$ is calculated by standard metric criterions (CIDEr, SPICE, etc.), the most important issue is the computation of discriminator probability output $p$.
One straightforward way to train a conditional GAN is to train the discriminator to judge pairs (image, sentence) as real or fake. This type of
conditioning is naive in the sense that the discriminator has no explicit notion of whether real training sentences match the extracted image features.

Typically, the discriminator observes three kinds of input pairs (one kind of positive samples and two kinds of negative samples):
ground-truth sentences with matched images (real pairs $(\bm{I},\bm{x}_{1:T})$), generated sentences with matched images (fake pairs $(\bm{I},\tilde{\bm{x}}_{1:T})$) and ground-truth sentences with mismatched images (wrong pairs $(\bm{I},\hat{\bm{x}}_{1:T})$). For the convenience of mathematical expression, we denote three sets as $\mathbb{S}_r, \mathbb{S}_f, \mathbb{S}_w$, which respectively consist of the three kinds of samples mentioned above.

Therefore, we slightly modify the traditional GAN training
loss function~\cite{GANs} to separate the two error sources. The model
parameters $\bm{\phi}$ of caption discriminator are trained to maximize
\begin{equation}\label{eq:discriminator_loss}
\begin{aligned}
L_D(\bm{\phi}) & = \mathbb{E}_{(\bm{I},\bm{x}_{1:T})\in\mathbb{S}_r}\left[\log D_{\bm{\phi}}(\bm{I},\bm{x}_{1:T})\right] \\
               & + 0.5\cdot\mathbb{E}_{(\bm{I},\tilde{\bm{x}}_{1:T})\in\mathbb{S}_f}\left[\log(1-D_{\bm{\phi}}(\bm{I},\tilde{\bm{x}}_{1:T}))\right]\\
               & + 0.5\cdot\mathbb{E}_{(\bm{I},\hat{\bm{x}}_{1:T})\in\mathbb{S}_w}\left[\log(1-D_{\bm{\phi}}(\bm{I},\hat{\bm{x}}_{1:T}))\right].\\
\end{aligned}
\end{equation}
Algorithm~\ref{alg::Image_Caption_GAN} describes the image captioning algorithm via the generative adversarial training method in detail. Notice that our proposed algorithm needs pre-training for both the generator and discriminator first. Then, we fine-tune the generator and discriminator alternatively based on the standard GAN training process.

\begin{algorithm}[ht]
\caption{Image Captioning Via Generative Adversarial Training Method}
\label{alg::Image_Caption_GAN}
\begin{algorithmic}[1]
\REQUIRE
caption generator $G_{\bm{\theta}}$;
discriminator $D_{\bm{\phi}}$;
language evaluator $Q$, e.g. CIDEr-D;
training set $\mathbb{S}_r=\{(\bm{I},\bm{x}_{1:T})\}$ and $\mathbb{S}_w=\{(\bm{I},\hat{\bm{x}}_{1:T})\}$.
\ENSURE
optimal parameters $\bm{\theta}$, $\bm{\phi}$.
\STATE Initial $G_{\bm{\theta}}$ and $D_{\bm{\phi}}$ randomly.
\STATE Pre-train $G_{\bm{\theta}}$ on $\mathbb{S}_r$ by MLE.
\STATE Generate some fake samples based on $G_{\bm{\theta}}$ to form $\mathbb{S}_f = \{(\bm{I}, \tilde{\bm{x}}_{1:T})\}$.
\STATE Pre-train $D_{\bm{\phi}}$ on $\mathbb{S}_r\cup\mathbb{S}_f\cup\mathbb{S}_w$ by Eq.~(\ref{eq:discriminator_loss}).
\REPEAT
    \FOR{g-steps=$1:g$}
        \STATE Generate a mini-batch of image-sentence pairs $\{(\bm{I},\tilde{\bm{x}}_{1:T})\}$ by $G_{\bm{\theta}}$.
        \STATE Calculate $p$ based on Eqs.~(\ref{eq:image_sentence_feature})-(\ref{eq:CNN_prob_output}) or Eqs.~(\ref{eq:RNN_Discriminator})-(\ref{eq:RNN_prob_output}).
        \STATE Calculate $s$ based on $Q$.
        \STATE Calculate reward $r$ according to Eq.~(\ref{eq:reward_function}).
        \STATE Update generator parameters $\bm{\theta}$ by SCST method via Eq.~(\ref{eq:RL_reward_gradient_sample_SCST}).
    \ENDFOR
    \FOR{d-steps=$1:d$}
        \STATE Generate negative image-sentence pairs $\{(\bm{I},\tilde{\bm{x}}_{1:T})\}$ by $G_{\bm{\theta}}$, together with negative samples $\{(\bm{I}, \hat{\bm{x}}_{1:T})\}\subseteq\mathbb{S}_w$ and positive samples $\{(\bm{I}, \bm{x}_{1:T})\}\subseteq\mathbb{S}_r$.
        \STATE Update discriminator parameters $\bm{\phi}$ via Eq.~(\ref{eq:discriminator_loss}).
    \ENDFOR
\UNTIL generator and discriminator converge
\end{algorithmic}
\end{algorithm}
\section{Experiments}\label{sec:experiments}
\subsection{Dataset}
The most widely used image captioning training and evaluation dataset is the MSCOCO dataset~\cite{MSCOCO} which contains 82,783, 40,504, and 40,775 images with 5 captions each for training,
validation, and test, respectively. For offline evaluation, following the Karpathy splits from~\cite{viual-semantic}, we use a set of 5K images for validation, 5K images for test and 113,287 images for training. We adopt five widely
used automatic evaluation metrics: BLEU, ROUGE-L,
METEOR, CIDEr and SPICE, to objectively evaluate the performance of different algorithms. For online evaluation on
the MSCOCO evaluation server, we add the 5K validation set and 5K testing set into the training set to
form a larger training set.

All the sentences in the training set are truncated to ensure the longest length of any sentence is 16 characters. We follow standard practice and perform some
text pre-processing, converting all sentences to lower case,
tokenizing on white space, and filtering words that do not
occur at least 5 times, resulting in a model vocabulary
of 9,487 words.

\subsection{Implementation Details}\label{subsec:implementation}
The LSTM hidden dimension for the RNN-based discriminator is 512. The dimension of image CNN feature and word embedding for both CNN-based and RNN-based discriminators is fixed to 2048.  We initialize the discriminator via pre-training the model for 10 epochs by minimizing the cross entropy loss in Eq.~(\ref{eq:discriminator_loss}) using the ADAM~\cite{ADAM} optimizer with a batch size of 16, an
initial learning rate of $1\times10^{-3}$ and momentum of 0.9 and 0.999. Similarly, the generator is also pre-trained by MLE for 25 epochs. We use a beam search with a beam
size of 5 when validating and tesing.
Notice that our proposed generative-adversarial-nets-based image captioning framework is generic so that any existing encoder-decoder model can be employed as the generator. In our experiments, the Resnet101~\cite{resnet} and bottom-up mechanism~\cite{image_caption_Top-Down} based on fastar R-CNN are chosen as encoders, the top-down attention model~\cite{image_caption_Top-Down}, att2in and att2all model~\cite{image_caption_self-critical} are chosen as decoders to conduct controlled experiments.
In addition, the SCST~\cite{image_caption_self-critical} reinforcement learning method is adopted by all experiments. Therefore, all the generator and reinforcement learning hyper-parameters are identical with those of original referenced papers.

Finally, the parameters that are unique and need to be determined in our algorithm setting are the balance factor $\lambda$ in Eq.~(\ref{eq:reward_function}), g-steps $g$ and d-steps $d$ during adversarial training in Algorithm~\ref{alg::Image_Caption_GAN}, and the language evaluator $Q$. All the above hyper-parameters are empirical values and will be clarified in the following subsection.
\subsection{Parameters Determination}
\begin{table}[t]
\centering
\caption{$\lambda$ selection} \label{tab:lambda_selection}
\resizebox{.99\columnwidth}{!}{
\begin{tabular}{c|c|c|c|c|c}
\toprule[2pt]
\multicolumn{6}{c}{fixed parameters: $g$=1; $d$=1; Metric=CIDEr-D}\\
\midrule[1pt]
$\lambda$         &   BLEU-4   &  METEOR    &  ROUGE-L     &   CIDEr   &   SPICE   \\
\midrule[1pt]
0                 &   0.363            &   0.277              &   0.569               &   1.201               &   0.214   \\
\textbf{0.3}      &   \textbf{0.383}   &   \textbf{0.286}     &   \textbf{0.586}      &   \textbf{1.232}      &   \textbf{0.221}  \\
0.5               &   0.368            &   0.285              &   0.581               &   1.215               &   0.220   \\
0.7               &   0.353            &   0.280              &   0.565               &   1.169               &   0.215   \\
1                 &   0.341            &   0.268              &   0.555               &   1.116               &   0.205   \\
\bottomrule[2pt]
\end{tabular}}
\end{table}

\begin{table*}[t]
\centering
\caption{Metric selection} \label{tab:metric_selection}
\begin{tabular}{c|c|c|c|c|c}
\toprule[2pt]
\multicolumn{6}{c}{fixed parameters: $g$=1; $d$=1; $\lambda$=0.3}\\
\midrule[1pt]
 Metric          &   BLEU-4   &  METEOR   &   ROUGE-L      &   CIDEr    &   SPICE   \\
\midrule[1pt]
CIDEr            &   0.381              &   0.280             &   0.580                 &   \textbf{1.248}     &   0.213   \\
\textbf{CIDEr-D}  &   \textbf{0.383}    &0.286      &   \textbf{0.586}                 &   1.232               &   \textbf{0.221}  \\
BLEU-4            &   0.383               &   0.279      &   0.574                      &   1.182              &   0.209  \\
ROUGE-L            &   0.368                &   0.283      &   0.585                    &   1.195              &   0.217  \\
METEOR            &   0.377               &   \textbf{0.287}      &   0.576              &   1.180                    &   0.214\\
\bottomrule[2pt]
\end{tabular}
\end{table*}

\begin{table*}[t]
\centering
\caption{Step size combination selection} \label{tab:stepsize_selection}
\begin{tabular}{c|c|c|c|c|c}
\toprule[2pt]
\multicolumn{6}{c}{fixed parameters:  $\lambda$=0.3; Metric=CIDEr-D}\\
\midrule[1pt]
 Step Sizes                   &   BLEU-4   &   METEOR   &   ROUGE-L          &   CIDEr   &   SPICE   \\
\midrule[1pt]
$g$=1; $d$=5                  &   0.378              &   0.284            &   0.582               &   1.209               &   0.220   \\
\textbf{$g$=1; $d$=1}         &   \textbf{0.383}      &   \textbf{0.286}   &   \textbf{0.586}     &   \textbf{1.232}      &   \textbf{0.221}  \\
$g$=5; $d$=1                  &   0.383               &   0.285           &   0.585               &   1.231               &   0.220  \\
$g$=10; $d$=1                 &   0.381                &   0.284           &   0.583              &   1.228               &   0.220  \\
\bottomrule[2pt]
\end{tabular}
\end{table*}

\begin{table*}[t]
\centering
\caption{Performance comparisons on MSCOCO Karpathy test set. The baseline algorithms are using resnet101 or bottom-up mechanism as the image feature extractor and SCST as the training method. Results of algorithms denoted by * are provided by original papers and the remaining experimental results are implemented by us for comparison. ``None'' means RL training method without discriminator. ``CNN-GAN'' and ``RNN-GAN'' mean training with our proposed approach by CNN-based and RNN-based discriminator, respectively. ``Ensemble'' indicates an ensemble of 4 CNN-GAN and 4 RNN-GAN models with different initializations. All values are reported in percentage (\%).} \label{tab:coco_test_comparison}
\resizebox{.99\textwidth}{!}{
\begin{tabular}{l|l|l|l|l|l|l|l|l|l|l|l}
\toprule[1.5pt]
Generator  &   Discriminator                    &   BLEU-1  & BLEU-2  & BLEU-3 & BLEU-4  &  METEOR    &   ROUGE-L   &CIDEr   &   SPICE  & CNN-D & RNN-D \\
\midrule[1pt]
                                & none*      &   -      &   -        &   -        &   33.3           &   26.3   &   55.3           &   111.4        &   -  &  -   & -  \\
resnet101+att2in                 &CNN-GAN    &  78.1      &   61.3     &   46.4     &   34.4           &   26.6   &   56.1           &   112.3        &   20.3 &  47.9   & 45.8  \\
\cite{image_caption_self-critical}&RNN-GAN   &  78.0      &   61.4     &   46.3     &   34.3           &   26.5   &   56.0           &   112.2        &   20.4 &  46.0   & 48.1  \\
                                 &\textbf{ensemble}   &   \textbf{78.5}      &   \textbf{61.8}    &   \textbf{47.1}    &   \textbf{35.0}    &   \textbf{27.1}   &   \textbf{56.6}      &   \textbf{114.8}        &   \textbf{20.5} &  \textbf{48.0}   & \textbf{48.2}  \\
\midrule[1pt]
                                & none        &   79.0      &   62.1        &   48.2        &   35.5         &   27.0   &   56.3          &   117.0     &   20.9  &  45.6   & 44.5  \\
bottom-up+att2in                &CNN-GAN      &   80.1      &   63.8        &   49.0        &   37.0         &   27.9   &   57.7          &   118.0     &   21.4 &  51.2   & 49.7  \\
\cite{image_caption_self-critical}&RNN-GAN    &   80.0      &   63.9        &   49.1        &   36.8         &   27.8   &   57.6          &   118.1     &   21.3 &  49.5   & 51.9  \\
                                 &\textbf{ensemble}    &   \textbf{80.5}      &   \textbf{64.8}    &   \textbf{50.0}    &   \textbf{37.9}    &   \textbf{28.4}   &   \textbf{58.2}      &   \textbf{119.5}        &   \textbf{21.5} &  \textbf{51.4}   & \textbf{51.5}  \\
\midrule[1pt]
                                 & none*      &   -          &   -        &   -         &   34.2       &   26.7     &    55.7     &   114.0     &         &  -   & -  \\
resnet101+att2all               &CNN-GAN      &   78.4      &   62.6    &   47.6     &   35.4       &   27.4     &    56.8     &   115.2     &   20.6 &  49.0   & 47.2  \\
\cite{image_caption_self-critical}&RNN-GAN    &   78.3      &   62.5    &   47.6     &   35.2       &   27.3     &    56.9     &   115.1     &   20.6 &  47.1   & 48.8  \\
                                 &\textbf{ensemble}    &   \textbf{79.0}      &   \textbf{62.8}    &   \textbf{48.2}    &   \textbf{35.8}    &   \textbf{27.7}   &   \textbf{57.6}      &   \textbf{117.8}        &   \textbf{20.9} &  \textbf{49.5}   & \textbf{49.1}  \\
\midrule[1pt]
                                 & none      &   79.6      &   63.5    &   49.1    &   36.1    &   27.8   &   56.7      &   119.8        &   21.2 &  46.3   & 45.9  \\
bottom-up+att2all               &CNN-GAN     &   80.7      &   64.7    &   50.1    &   38.0    &   28.4   &   58.4      &   122.1        &   21.9 &  53.5   & 50.8  \\
\cite{image_caption_self-critical}&RNN-GAN   &   80.6      &   64.8    &   50.0    &   38.1    &   28.3   &   58.3      &   122.0        &   21.8 &  50.6   & 53.2  \\
                                 &\textbf{ensemble}   &   \textbf{81.1}      &   \textbf{65.7}    &   \textbf{50.8}    &   \textbf{39.0}    &   \textbf{28.6}   &   \textbf{58.7}      &   \textbf{124.1}        &   \textbf{22.0} &  \textbf{53.7}   & \textbf{53.5}  \\
\midrule[1pt]
                             & none*      &   76.6      &   -        &   -        &   34.0    &   26.5   &   54.9      &   111.1        &   20.2 &  -   & -  \\
resnet101+top-down           &CNN-GAN     &   78.5      &   62.7    &   48.0    &   35.6    &   27.3   &   56.7      &   113.0        &   20.6 &  49.5   & 47.6  \\
\cite{image_caption_Top-Down}&RNN-GAN     &   78.4      &   62.7    &   48.0    &   35.5    &   27.2   &   56.6      &   112.7        &   20.5 &  47.0   & 49.2  \\
                             &\textbf{ensemble}    &   \textbf{79.3}      &   \textbf{63.2}    &   \textbf{48.6}    &   \textbf{36.0}    &   \textbf{27.6}   &   \textbf{57.1}      &   \textbf{115.5}        &   \textbf{20.8} &  \textbf{50.0}   & \textbf{49.3}  \\
\midrule[1pt]
                                & none*      &   79.8      &   -        &   -        &   36.3    &   27.7   &   56.9      &   120.1        &   21.4 &  -   & -  \\
bottom-up+top-down              &CNN-GAN     &   81.1      &   65.0    &   50.4    &   38.3    &   28.6   &   58.6      &   123.2        &   22.1 &  53.6   & 51.1  \\
\cite{image_caption_Top-Down}   &RNN-GAN     &   81.0      &   64.8    &   50.2    &   38.2    &   28.5   &   58.4      &   122.2        &   22.0 &  50.9   & 54.0  \\
                                 &\textbf{ensemble}   &   \textbf{81.8}      &   \textbf{66.1}    &   \textbf{51.6}    &   \textbf{39.6}    &   \textbf{28.9}   &   \textbf{59.1}      &   \textbf{125.9}        &   \textbf{22.3} &  \textbf{54.3}   & \textbf{54.5}  \\
\midrule[1pt]
\textbf{Average}    &  \textbf{CNN-GAN}  &   \textbf{1.71}   &   \textbf{2.31}     &   \textbf{1.85}      &   \textbf{4.44}      &   \textbf{2.59}  &
                                                    \textbf{2.53}      &   \textbf{1.50}      &   \textbf{2.75} &  \textbf{13.93}   & \textbf{11.17}  \\
\textbf{Improvements}  &  \textbf{RNN-GAN}  &   \textbf{1.59}   &   \textbf{2.47}     &   \textbf{1.85}      &   \textbf{4.15}      &   \textbf{2.22}  &   \textbf{2.38}      &   \textbf{1.28}      &   \textbf{2.27} &  \textbf{8.92}   & \textbf{16.26}  \\
\bottomrule[1.5pt]
\end{tabular}}
\end{table*}

\begin{table*}[!ht]
\centering
\caption{Performance of different models on the MSCOCO evaluation server. All values are reported in percentage (\%), with the highest value of each entry highlighted in boldface. It is worth pointing out that almost all the metrics of our method (ensemble of 4 CNN-GAN and 4 RNN-GAN models in the last row) ranked in top two at the time of submission (5 Sep., 2018).} \label{tab:coco_online_comparison}
\resizebox{.99\textwidth}{!}{
\begin{tabular}{l|l|l|l|l|l|l|l|l|l|l|l|l|l|l|l|l}
\toprule[1.5pt]
\multirow{2}{*}{Algorithms}  &\multicolumn{2}{c|}{BLEU-1}&\multicolumn{2}{c|}{BLEU-2}&\multicolumn{2}{c|}{BLEU-3}&\multicolumn{2}{c|}{BLEU-4}&\multicolumn{2}{c|}{METEOR}&\multicolumn{2}{c|}{ROUGE-L}&\multicolumn{2}{c|}{CIDEr}&
\multicolumn{2}{c}{SPICE}\\
\cmidrule[0.5pt](lr){2-3}\cmidrule[0.5pt](lr){4-5}\cmidrule[0.5pt](lr){6-7}\cmidrule[0.5pt](lr){8-9}\cmidrule[0.5pt](lr){10-11}\cmidrule[0.5pt](lr){12-13}\cmidrule[0.5pt](lr){14-15}
\cmidrule[0.5pt](lr){16-17}
& c5     &  c40    & c5     &  c40  & c5     &  c40  & c5     &  c40  & c5     &  c40  & c5     &  c40  & c5     &  c40  & c5     &  c40  \\
\midrule[1pt]
NIC~\cite{show_tell}                             & 71.3   &  89.5     &54.2    &80.2   &40.7    &  69.4 & 30.9   &58.7   &25.4    &34.6   &53.0    &68.2   & 94.3   & 94.6  &  18.2 & 63.6 \\
PG-BCMR~\cite{ImageCaptionPolicyGradientSpider}  & 75.4   &  91.8     &59.1    &84.1   &44.5    &  73.8 & 33.2   &62.4   &25.7    &34.0   &55.0    &69.5   &101.3   &103.2  &  18.7 & 62.2 \\
Adaptive~\cite{image_caption_adaatt}             & 74.8   &  92.0     &58.4    &84.5   &44.4    &  74.4 & 33.6   &63.7   &26.4    &35.9   &55.0    &70.5   &104.2   &105.9  &  19.7 & 67.3 \\
Actor-Critic~\cite{ImageCaptionActorCritic}      & 77.8   &  92.9     &61.2    &85.5   &45.9    &  74.5 & 33.7   &62.5   &26.4    &34.4   &55.4    &69.1   &110.2   &112.1  &  20.3 & 68.0 \\
Att2all~\cite{image_caption_self-critical}       & 78.1   &  93.7     &61.9    &86.0   &47.0    &  75.9 & 35.2   &64.5   &27.0    &35.5   &56.3    &70.7   &114.7   &116.7  &  20.7 & 68.9 \\
Stack-Cap~\cite{Stack-Cap}                       & 77.8   &  93.2     &61.6    &86.1   &46.8    &  76.0 & 34.9   &64.6   &27.0    &35.6   &56.2    &70.6   &114.8   &118.3  &  -    &  -  \\
LSTM-A$_3$~\cite{LSTMA3}                         & 78.7   &  93.7     &62.7    &86.7   &47.6    &  76.5 & 35.6   &65.2   &27.0    &35.4   &56.4    &70.5   &116.0   &118.0  & - & - \\
Up-down~\cite{image_caption_Top-Down}            & 80.2   &  95.2     &64.1    &88.8   &49.1    &  79.4 & 36.9   &68.5   &27.6    &36.7   &57.1    &72.4   &117.9   &120.5  &  21.5 & 71.5 \\
CAVP~\cite{ContextAwareVP}                       & 80.1   &  94.9     &64.7    &88.8   &50.0    &  79.7 & 37.9   &69.0   &28.1    &37.0   &58.2    &73.1   &121.6   &123.8  &  -     & -  \\
RFNet~\cite{RecurrentFusion}                     & 80.4   &  95.0     &64.9    &89.3   &50.1    &  80.1 & 38.0   &69.2   &28.2    &37.2   &58.2    &73.1   &122.9   &\textbf{125.1}  &  -     & -  \\
\midrule[1pt]
Ours                           & \textbf{81.9}   &  \textbf{95.6}     &\textbf{66.3}    &\textbf{90.1}   &\textbf{51.7}    &  \textbf{81.7} & \textbf{39.6}   &\textbf{71.5}   &\textbf{28.7}    &\textbf{38.2}  &\textbf{59.0}    &\textbf{74.4}   &\textbf{123.1}   &124.3  &  -     & -  \\
\bottomrule[1.5pt]
\end{tabular}}
\end{table*}

\begin{table*}[!ht]
\scriptsize
\centering
\caption{This table shows some caption examples by different algorithms. The first line is generated by the Up-Down model with SCST. The second and third lines are caption results by adding CNN-GAN and RNN-GAN traing, respectively. The last line is the results of ensemble of CNN-GAN and RNN-GAN.} \label{tab:example}
\resizebox{.99\textwidth}{!}{
\begin{tabular}{|l|l|l|l|l|}
\hline
&\includegraphics[width=3.6cm,height=2.8cm]{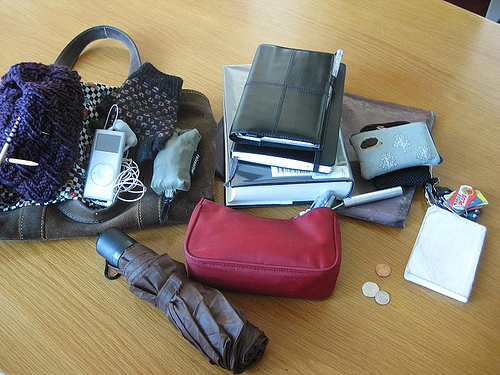}  &   \includegraphics[width=3.6cm,height=2.8cm]{COCO_val2014_000000307292}  &   \includegraphics[width=3.6cm,height=2.8cm]{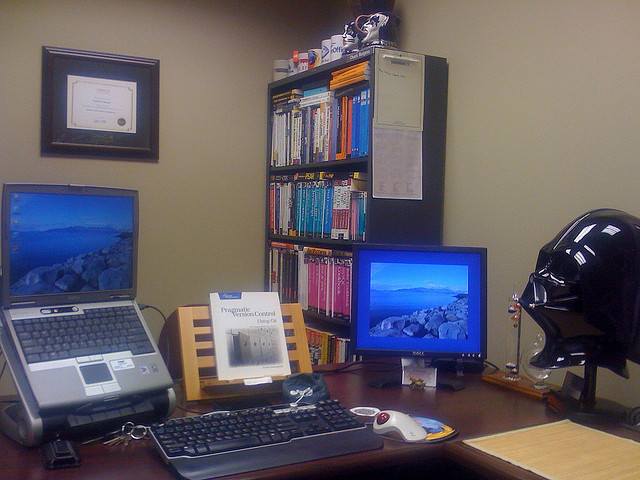} & \includegraphics[width=3.6cm,height=2.8cm]{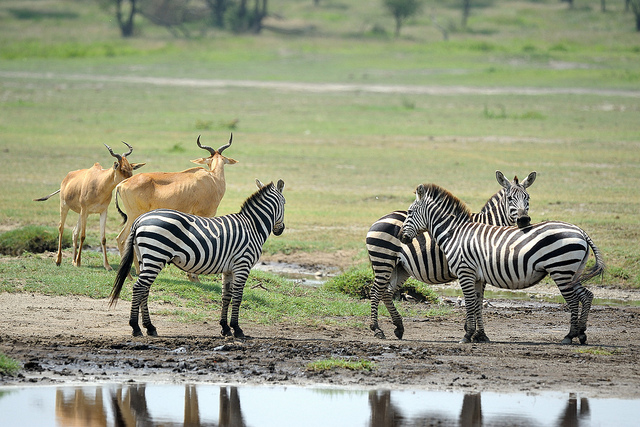}\\
\hline
\multirow{2}{*}{Up-Down} &a bag of \textbf{items and other items} &a man with a beard and \textbf{tie wearing} &a desk with \textbf{two laptops and a}&a group of \textbf{zebras and a zebra}    \\
        &   on a table                                       & \textbf{a tie}                          &  \textbf{laptop}                 &standing in the water\\
\hline
\multirow{2}{*}{CNN-GAN} &a \textbf{purse and personal items} laid out &a man in a suit and tie \textbf{looking at}   &a desk with \textbf{a laptop computer and}&a group of \textbf{zebras and other }       \\
        & on a wooden table& \textbf{the camera}& \textbf{a desktop} on it              &\textbf{animals} in the water\\
\hline
\multirow{2}{*}{RNN-GAN} &a \textbf{purse and other items} laid out on   &a man in a suit and tie \textbf{is smiling}  &a desk with \textbf{a laptop computer and}&a group of \textbf{zebras and other}\\
        &a wooden table & &\textbf{books} on it            & \textbf{animals} standing in the water\\
\hline
\multirow{2}{*}{ensemble} &a \textbf{purse and other personal items}  &a man in a suit and tie \textbf{looking at the}          &a desk with \textbf{a laptop and a desktop}    &a group of \textbf{zebras and other}\\
         &  laid out on a wooden table                               &\textbf{camera with smile}                  &      sitting on top of it           &\textbf{animals} standing near the water \\
\midrule[1pt]
&\includegraphics[width=3.6cm,height=2.8cm]{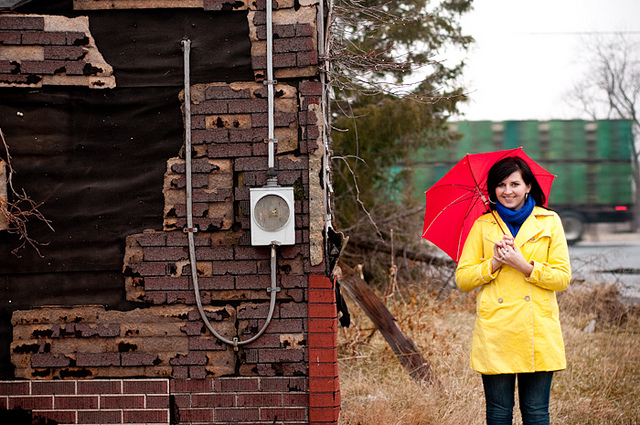}  &   \includegraphics[width=3.6cm,height=2.8cm]{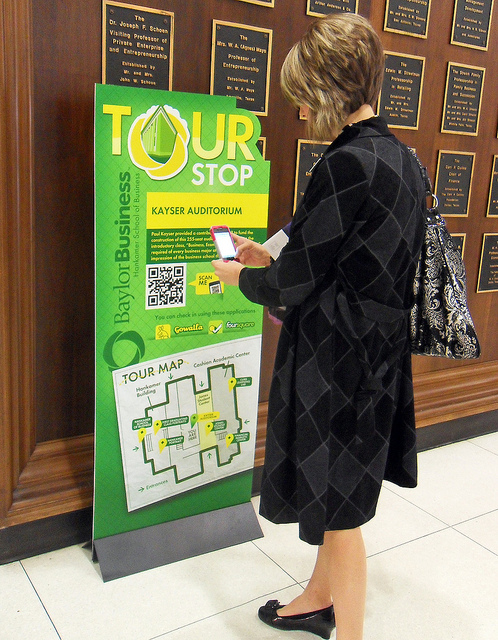}  &   \includegraphics[width=3.6cm,height=2.8cm]{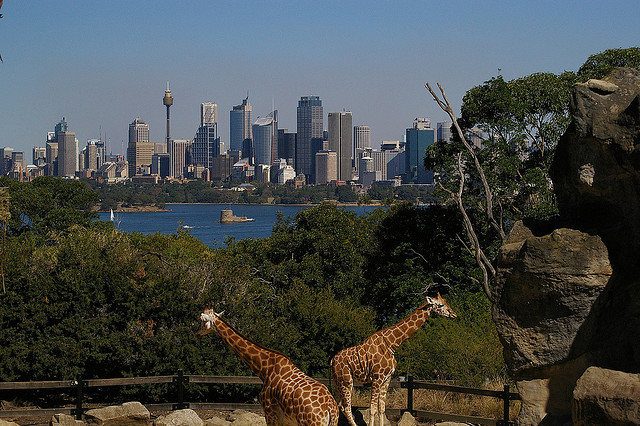}  & \includegraphics[width=3.6cm,height=2.8cm]{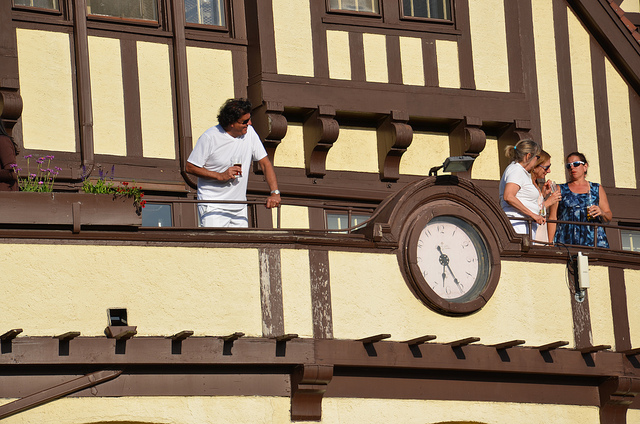}\\
\hline
\multirow{2}{*}{Up-Down}   &a woman holding an umbrella  &a woman standing \textbf{in front of a}&two giraffes standing next to a  &a group of people standing \textbf{on top of}    \\
        & \textbf{in a brick wall}&  \textbf{cell phone}  &\textbf{city in the water}&\textbf{a clock}\\
\hline
\multirow{2}{*}{CNN-GAN}   &a woman in a yellow jacket    &a woman standing \textbf{in front of a}     &two giraffes standing next to a & a group of people standing \textbf{on a} \\
    & holding an umbrella& \textbf{newspaper sign}&large \textbf{city in the background}& \textbf{building with a clock}\\
\hline
\multirow{2}{*}{RNN-GAN}   &a woman standing \textbf{in front of} &a woman standing \textbf{in front of a}   &two giraffes standing next to a  &a group of people standing \textbf{on a} \\
          & \textbf{a brick wall} holding an umbrella                                         & \textbf{store} holding a cell phone                                          &   \textbf{city in the background}   &\textbf{balcony looking at a clock}                                         \\
\hline
\multirow{2}{*}{ensemble}   &a woman in a yellow jacket \textbf{near}  &a woman standing \textbf{in front of a new-}  &two giraffes standing next to a   & a group of people standing \textbf{on a}\\
           &\textbf{a brick wall} holding an umbrella            & \textbf{spaper sign} holding a cell phone         &   \textbf{city near water in the background}   & \textbf{balcony with a clock}                                     \\
\hline
\end{tabular}}
\end{table*}

In order to determine a group of optimal hyper-parameters as mentioned in the above subsection, we design a series of experiments with a variable-controlling approach. We adopt the bottom-up image feature extractor together with top-down attention model as our fixed generator, SCST as our RL optimization method and the CNN structure as our discriminator.

First, we fix the g-steps $g=1$, d-steps $d=1$ and language metric $Q$ as CIDEr-D (a smooth modification version of CIDEr). The objective results on the test split with different $\lambda$ values are shown in Table~\ref{tab:lambda_selection}. Notice that when $\lambda=0$, our algorithm exactly degenerates into the conventional RL method. Statistics reveal that all the metrics evaluation results achieve their optimal scores when $\lambda=0.3$.

Second, we fix the g-steps $g=1$, d-steps $d=1$ and $\lambda=0.3$. The test results while RL optimizing by different language evaluator $Q$ are shown in Table~\ref{tab:metric_selection}. Notice that here we do not choose SPICE as evaluator since the computation of SPICE is extremely slow and is too time-consuming. When the evaluator is chosen as CIDEr-D, even though some scores (CIDEr and METEOR) are not the highest, the comprehensive performance still outperforms other evaluators.

Third, we fix $\lambda=0.3$ and language evaluator $Q$ as CIDEr-D. We try different step-size combinations and list the test results in Table~\ref{tab:stepsize_selection}. Experimental results demonstrate that the best step-size combination is $g=1$ and $d=1$.

Overall, based on the experimental results explained above, the final optimal hyper-parameters of our proposed algorithm are $\lambda=0.3$, $g=1$, $d=1$ and $Q=$ CIDEr-D.
\subsection{Comparisons}
We compare our framework with some state-of-the-art encoder-decoder models (att2in~\cite{image_caption_self-critical}, att2all~\cite{image_caption_self-critical} and top-down attention~\cite{image_caption_Top-Down}) with SCST~\cite{image_caption_self-critical} to study the effectiveness of our proposed algorithm. For fair comparisons, we chose Resnet101~\cite{resnet} and bottom-up mechanism~\cite{image_caption_Top-Down} as the CNN feature extractors for all the models mentioned above and the identical decoder parameters as reported in the original papers. Statistics reported in Table~\ref{tab:coco_test_comparison} reveal that by using our proposed generative adversarial training method, the performance of all the objective evaluation metrics is improved for all three models. Specifically, the relative improvements range from  1.3\% to 4.4\% on average. Notice that the well-trained CNN-based and RNN-based discriminators (CNN-D and RNN-D in the last two columns of Fig.~\ref{tab:coco_test_comparison}) can also be viewed as two learned evaluators. The experimental results demonstrate that our proposed adversarial training approach can significantly boost the scores compared with traditional RL training method (improvements range from 8.9\% to 16.3\%).

In terms of CNN-based and RNN-based generative adversarial training framework (called CNN-GAN and RNN-GAN in Table~\ref{tab:coco_test_comparison}), each has its own advantages. Experimental results show that CNN-GAN can improve the performance of image captioning frameworks slightly more as compared with RNN-GAN. However, during training stage, using RNN-GAN can save 30\% training time according to our experimental experience. The most important issue is that the ensemble results of 4 CNN-GAN and 4 RNN-GAN models can largely enhance the performance of a single model as shown in Table~\ref{tab:coco_test_comparison}.

For online evaluation, we use the ensemble of 4 CNN-GAN and 4 RNN-GAN models with different random initializations whose generator structure exactly follows the Up-Down model~\cite{image_caption_Top-Down} and the comparisons are provided in Table~\ref{tab:coco_online_comparison}. We can see that our approach achieves the best results compared with some state-of-the-art algorithms with publicly published papers. For all submissions online, almost all the metrics of our method ranked in top two at the time of submission (5 Sep., 2018). Specifically, when compared with the results of Up-Down model, by using our proposed generative adversarial training method, it can obtain significant improvements.

\subsection{Examples}
To better understand the CNN-GAN and RNN-GAN framework, Table~\ref{tab:example} provides some representative examples for comparisons. The baseline algorithm is the Up-Down model with SCST, whose results are shown in the first line. Our generated captions by CNN-GAN and RNN-GAN training approach with the same generator are listed in the second and third line, respectively. The last row is the results of ensemble of 4 CNN-GAN and 4 RNN-GAN models.

The four cases appearing above in Table~\ref{tab:example} indicate that the original model will generate some duplicate words or phrase. Even without any grammar or description errors, these sentences seem rigid and machine generated. After generative adversarial training, the generated sentences are much more like a human style. The four cases appearing below in Table~\ref{tab:example} present some logical errors which are inconsistent with given images when using traditional RL training method, \textit{e.g.} ``in a brick wall'', ``in front of a cell phone'', ``city in the water'', ``on top of a clock'', while such error will be avoided when employing our proposed method.

\section{Conclusion}\label{sec:conclusion}
This paper proposes a novel architecture combining generative adversarial nets and reinforcement learning to improve existing image captioning frameworks. Current RL-based image captioning algorithms directly optimize language evaluation metrics such as CIDEr, BELU and SPICE; however, simply optimizing one metric or combination of these metrics will not consistently improve all evaluation metrics and will also result in some logical errors or unnatural styles when generating descriptions. Therefore, we are motivated to design a discriminator network to judge whether the input sentence is human described or machine generated. Alternatively, the caption generator and the evaluation discriminator are improved by adversarial training.
We experimentally show consistent improvements over all language evaluation metrics for different state-of-the-art encoder-decoder based image captioning models optimized by conventional RL training, demonstrating the effectiveness of our proposed generative adversarial training method. In addition, the well-trained CNN and RNN-based discriminators can also be utilized as image captioning evaluators leaned by neural networks.

{\fontsize{9pt}{10pt} \selectfont
\bibliographystyle{aaai}
\bibliography{egbib}
}

\section*{Supplementary Materials}
To better understand the CNN-GAN and RNN-GAN framework, Tables~\ref{tab:example1} and \ref{tab:example2} provide some representative examples for comparisons. The baseline algorithm is the Up-Down model with SCST, whose results are shown in the first line. Our generated captions by CNN-GAN and RNN-GAN training approach with the same generator are listed in the second and third line, respectively. The last row is the results of ensemble of 4 CNN-GAN and 4 RNN-GAN models.

The cases appearing in Table~\ref{tab:example1} indicate that the original model will generate some duplicate words or phrase such as ``skis and skis'', ``tow laptops and a laptop'', ``a bird and a bird'' and ``beach near the beach''. Even without any grammar or description errors, these sentences seem rigid and machine generated. After generative adversarial training, the generated sentences are much more like a human style.

The cases appearing in Table~\ref{tab:example2} present some logical errors which are inconsistent with given images when using traditional RL training method, \textit{e.g.} ``in front of a basketball'', ``a giraffe with its wings'', ``a beach in the water'', ``girls on a wall'', while such error will be avoided when employing our proposed method.

\begin{figure*}[ht]
  \centering
  \centerline{\includegraphics[width=17cm]{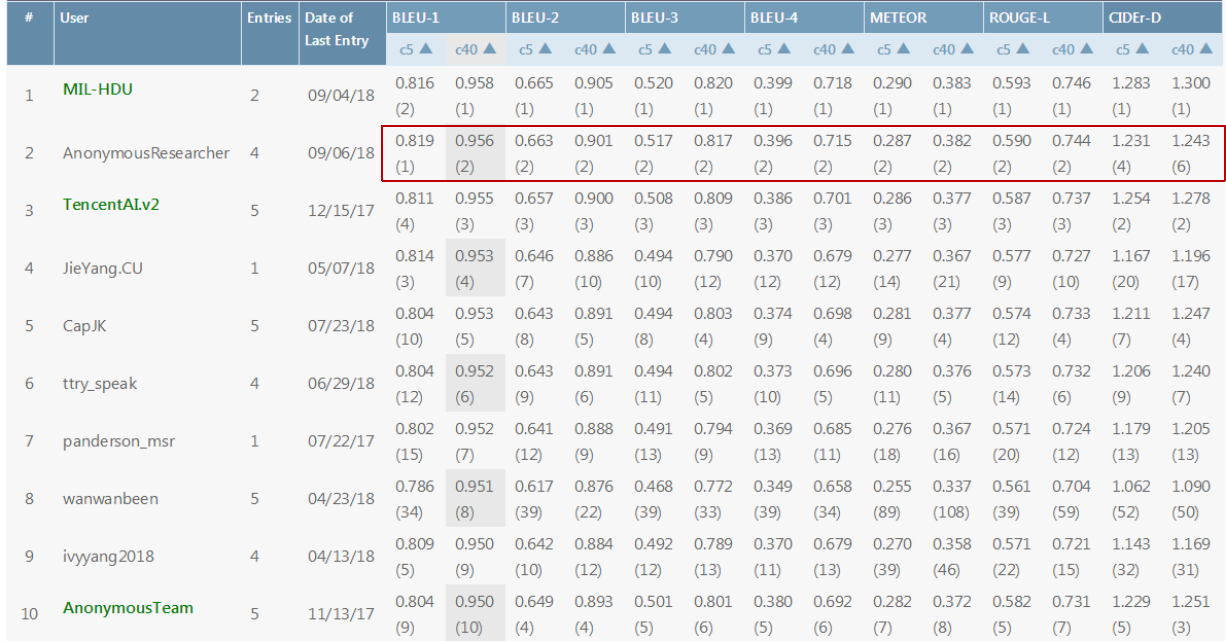}}
\caption{The results and rankings of state-of-the-art algorithms submitted to MSCOCO online test server. Notice that the scores of our submitted for each entry almost rank in top two at the date of submission (5 Sep., 2018).}
\label{fig:ranking}
\end{figure*}

Fig.~\ref{fig:ranking} shows the results and rankings of state-of-the-art algorithms submitted to MSCOCO online test server (\url{https://competitions.codalab.org/competitions/3221\#results}). Notice that the scores of our submitted method (ensemble of 4 CNN-GAN and 4 RNN-GAN models) for each entry almost rank in top two at the date of submission (5 Sep., 2018).

\begin{table*}[!ht]
\renewcommand\arraystretch{0.8}
\scriptsize
\begin{center}
\setlength{\tabcolsep}{0.1mm}
\caption{This table shows some caption examples by different algorithms. The first line is generated by the Up-Down model with SCST. The second and third lines are caption results by adding CNN-GAN and RNN-GAN traing, respectively. The last line is the results of ensemble of CNN-GAN and RNN-GAN.} \label{tab:example1}
\resizebox{.99\textwidth}{!}{
\begin{tabular}{|l|l|l|l|l|}
\hline
&\includegraphics[width=4.1cm,height=2.8cm]{COCO_val2014_000000232173}  &   \includegraphics[width=4.1cm,height=2.8cm]{COCO_val2014_000000085329}  &   \includegraphics[width=4.1cm,height=2.8cm]{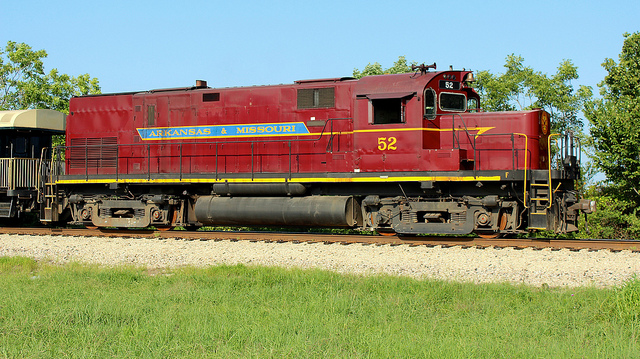} & \includegraphics[width=4.1cm,height=2.8cm]{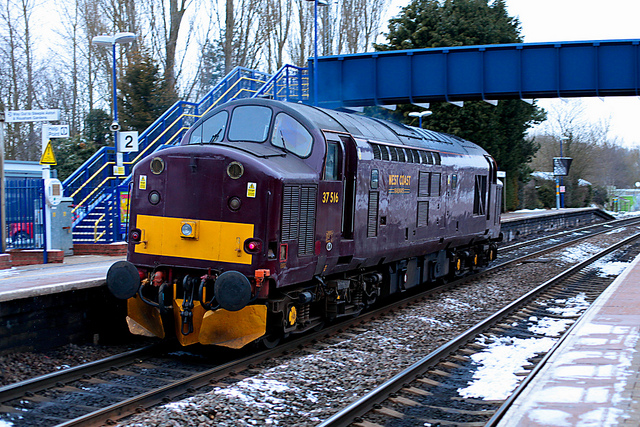}\\
\hline
\multirow{2}{*}{Up-Down}
&a pair of \textcolor{red}{skis and skis} in the floor &a woman with \textcolor{red}{a hair and her hair} &a red train is \textcolor{red}{on the tracks of a track}&a train is \textcolor{red}{on the tracks in a track}   \\
&                                                       &  in the air                                    &                                                      &\\
\hline
\multirow{2}{*}{CNN-GAN}
&a pair of \textcolor{blue}{skis} sitting on the floor &a woman with \textcolor{blue}{a white shirt and a tie}   &a red and yellow train is sitting&a purple and yellow train is \textcolor{blue}{on the tracks}      \\
& in a bathroom                     &                                                           &  \textcolor{blue}{on the tracks}               &\\
\hline
\multirow{2}{*}{RNN-GAN}
&a pair of \textcolor{blue}{skis and poles} sitting   &a close up of a woman \textcolor{blue}{wearing a tie}  &a large red train is sitting &a purple train \textcolor{blue}{on the tracks under a bridge} \\
& on a tile floor                                       &                                   &\textcolor{blue}{on the tracks}                                                  &\\
\hline
\multirow{2}{*}{ensemble}
&a pair of \textcolor{blue}{skis and poles} on a tile floor &a woman wearing \textcolor{blue}{a white shirt and a tie}&a red and yellow train is \textcolor{blue}{on the tracks}&a purple and yellow train is \textcolor{blue}{on the tracks}\\
&                                                           &                                                           &                                    & \textcolor{blue}{under a bridge}  \\
\midrule[1pt]
&\includegraphics[width=4.1cm,height=2.8cm]{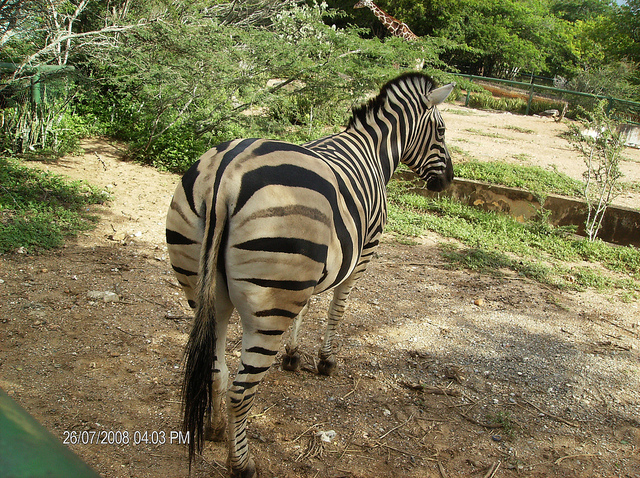}  &   \includegraphics[width=4.1cm,height=2.8cm]{COCO_val2014_000000281376}  &   \includegraphics[width=4.1cm,height=2.8cm]{COCO_val2014_000000520482}  & \includegraphics[width=4.1cm,height=2.8cm]{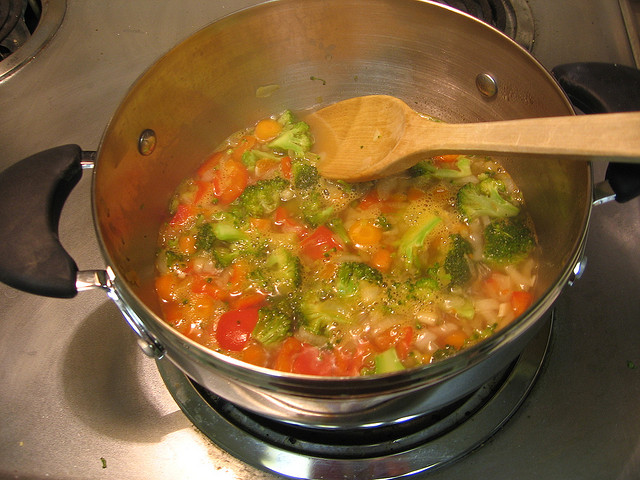}\\
\hline
\multirow{2}{*}{Up-Down}
&a zebra standing \textcolor{red}{in the dirt of a dirt} &a desk with \textcolor{red}{two laptops and a laptop}&\textcolor{red}{a bird and a bird} in a fence&a pot of food with \textcolor{red}{vegetables} \\
&                                                        &computer  &                           &\textcolor{red}{and vegetables} in a stove\\
\hline
\multirow{2}{*}{CNN-GAN}
&a zebra standing \textcolor{blue}{next to a dirt road}  &a desk with \textcolor{blue}{two computer monitors and}& \textcolor{blue}{a large bird} standing next to a fence&a pot of food with \textcolor{blue}{broccoli} \\
&                                                         & \textcolor{blue}{a laptop} on it&                           & \textcolor{blue}{and vegetables} in a stove\\
\hline
\multirow{2}{*}{RNN-GAN}
&a zebra standing \textcolor{blue}{on the side of a dirt road} &a desk with \textcolor{blue}{two monitors and a laptop}&\textcolor{blue}{two large birds} are standing in a cage&a pot filled with \textcolor{blue}{vegetables} in a stove \\
&                                                               &  computer on it                    &               &                                       \\
\hline
\multirow{2}{*}{ensemble}
&a zebra standing \textcolor{blue}{on a dirt road}  &a desk with \textcolor{blue}{two monitors and a laptop}&\textcolor{blue}{a couple of birds} standing in a cage&a pot of food with \textcolor{blue}{broccoli}\\
&                                                   &  computer on it          &                      &  \textcolor{blue}{and vegetables} in a stove  \\
\midrule[1pt]
&\includegraphics[width=4.1cm,height=2.8cm]{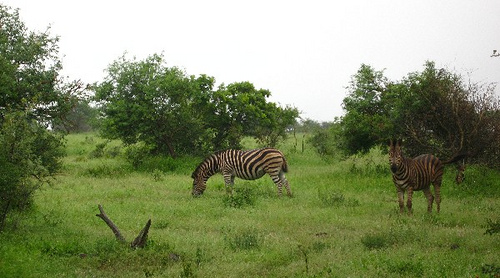}  &   \includegraphics[width=4.1cm,height=2.8cm]{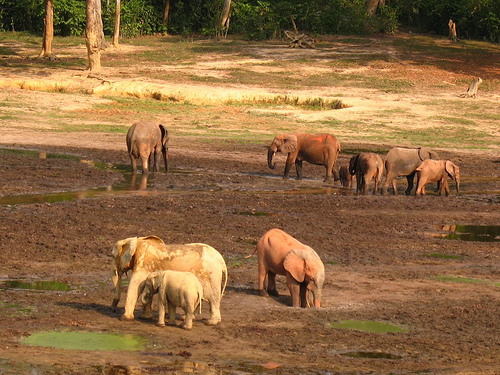}  &   \includegraphics[width=4.1cm,height=2.8cm]{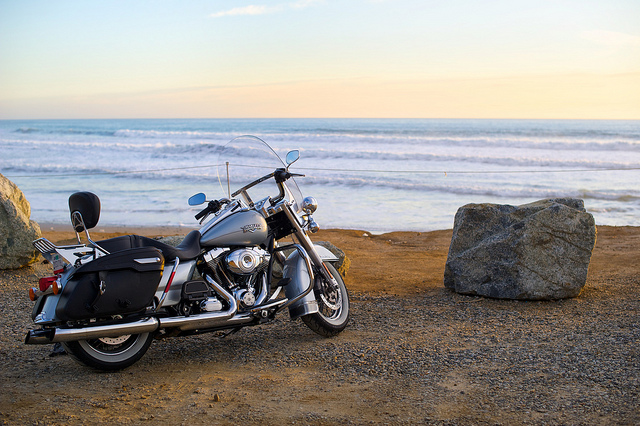}  & \includegraphics[width=4.1cm,height=2.8cm]{COCO_val2014_000000535262}\\
\hline
\multirow{2}{*}{Up-Down}
&\textcolor{red}{two zebras and a zebra} grazing in a field  &a herd of \textcolor{red}{elephants and elephants} in a &a motorcycle parked on the&an elephant walking in the \\
&                                                        &field                                             &  \textcolor{red}{beach near a beach}  &\textcolor{red}{dirt of a dirt road}\\
\hline
\multirow{2}{*}{CNN-GAN}
&\textcolor{blue}{two zebras} grazing in a field of grass   &a herd of elephants walking in a field&a motorcycle parked on the&an elephant walking through \textcolor{blue}{a dirt field} \\
&                                                         &                                     &    \textcolor{blue}{side of a beach}    & \\
\hline
\multirow{2}{*}{RNN-GAN}
&\textcolor{blue}{two zebras} grazing in a field of green   &a herd of elephants walking in a field &a motorcycle parked on a&an elephant walking across  \\
&grass                                                  & \textcolor{blue}{with other animals}   &\textcolor{blue}{beach near the ocean}   & \textcolor{blue}{a dirt covered field}      \\
\hline
\multirow{2}{*}{ensemble}
&\textcolor{blue}{two zebras} grazing in a field of grass  &a herd of elephants walking in a field&a motorcycle parked on the&an elephant walking in \textcolor{blue}{ a dirt field }\\
&                                                   &   \textcolor{blue}{with other animals}           &   \textcolor{blue}{side of a beach}        &    \\
\hline
\end{tabular}}
\end{center}
\end{table*}

\begin{table*}[!ht]
\renewcommand\arraystretch{0.8}
\scriptsize
\begin{center}
\setlength{\tabcolsep}{0.1mm}
\caption{This table shows some caption examples by different algorithms. The first line is generated by the Up-Down model with SCST. The second and third lines are caption results by adding CNN-GAN and RNN-GAN traing, respectively. The last line is the results of ensemble of CNN-GAN and RNN-GAN.} \label{tab:example2}
\resizebox{.99\textwidth}{!}{
\begin{tabular}{|l|l|l|l|l|}
\hline
&\includegraphics[width=4.1cm,height=2.8cm]{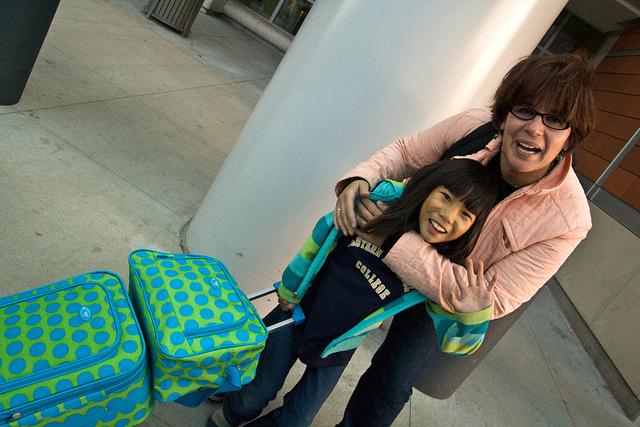}  &   \includegraphics[width=4.1cm,height=2.8cm]{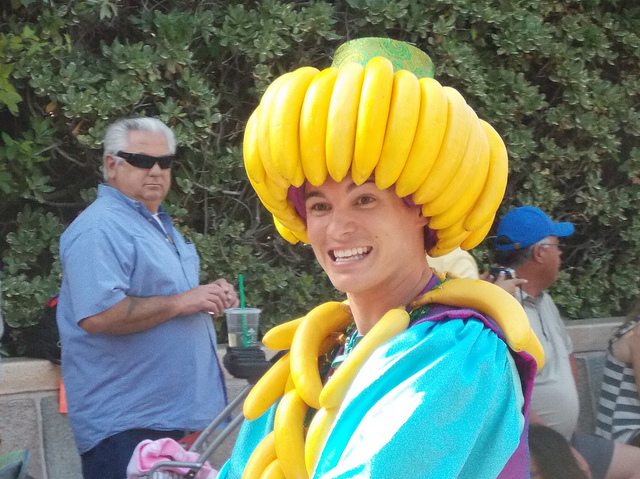}  &   \includegraphics[width=4.1cm,height=2.8cm]{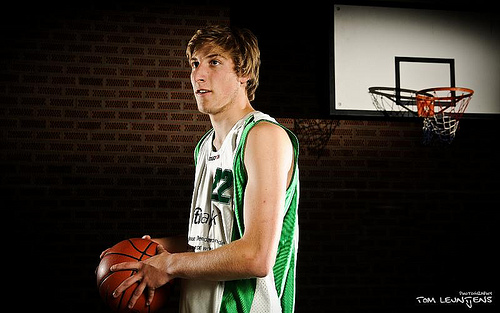} & \includegraphics[width=4.1cm,height=2.8cm]{COCO_val2014_000000527248}\\
\hline
\multirow{2}{*}{Up-Down}
&a woman  \textcolor{red}{holding a suitcase} &a man and a woman \textcolor{red}{standing} &a man standing \textcolor{red}{in front of a}&a crowd of people standing \textcolor{red}{in a building }   \\
&in the airport                               &\textcolor{red}{next to a banana}           &\textcolor{red}{basketball ball}             &\\
\hline
\multirow{2}{*}{CNN-GAN}
&a woman with glasses is \textcolor{blue}{sitting on the} &a man with a yellow banana \textcolor{blue}{in front} &a young man holding a basketball&a crowd of people standing \textcolor{blue}{in front}  \\
& \textcolor{blue}{ground with luggage}                   &\textcolor{blue}{of his head}                          &\textcolor{blue}{in his hand}                 &\textcolor{blue}{of a building} \\
\hline
\multirow{2}{*}{RNN-GAN}
&a woman and a child \textcolor{blue}{sitting on the}  &a man with a bunch of bananas  &a young man holding a basketball&a crowd of people standing \textcolor{blue}{in front} \\
& \textcolor{blue}{ground with luggage}                & \textcolor{blue}{on his head}    &\textcolor{blue}{in his hand}       &\textcolor{blue}{of a church}\\
\hline
\multirow{2}{*}{ensemble}
&a woman and a child \textcolor{blue}{sitting next to} &a man with a bunch of bananas&a young man holding a basketball&a crowd of people standing \textcolor{blue}{in front}\\
& \textcolor{blue}{a pile of luggage}                  &   \textcolor{blue}{on his head}   & \textcolor{blue}{in his hand}& \textcolor{blue}{of a building}   \\
\midrule[1pt]
&\includegraphics[width=4.1cm,height=2.8cm]{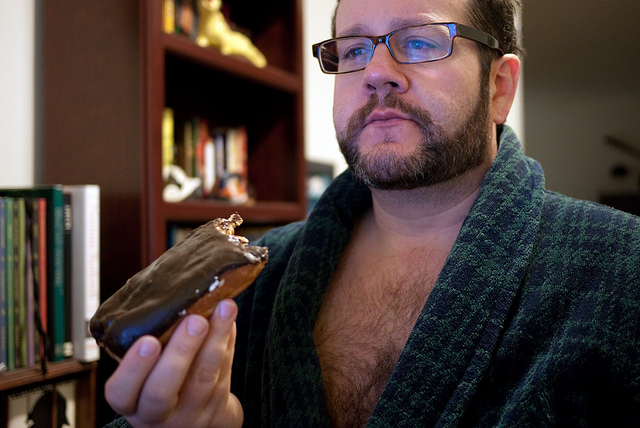}  &   \includegraphics[width=4.1cm,height=2.8cm]{COCO_val2014_000000068203}  &   \includegraphics[width=4.1cm,height=2.8cm]{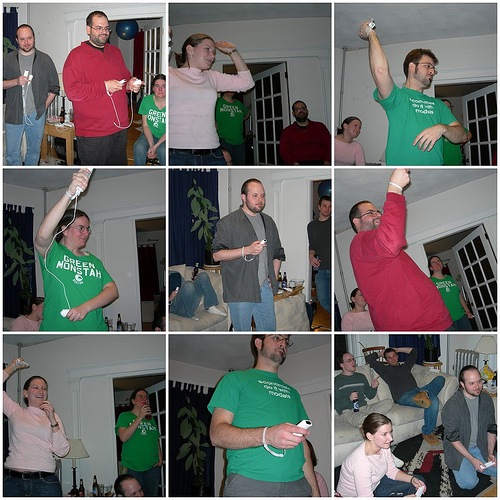}  & \includegraphics[width=4.1cm,height=2.8cm]{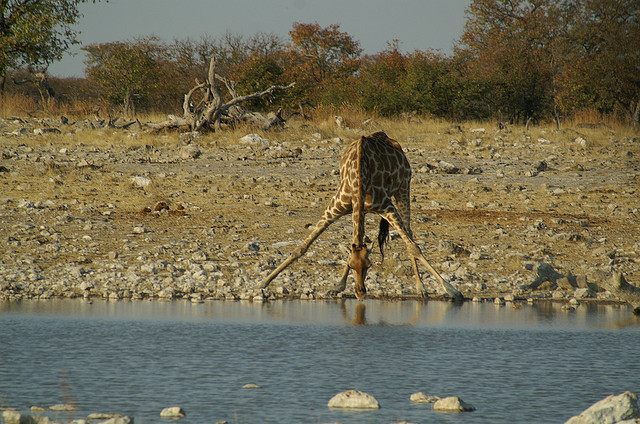}\\
\hline
\multirow{2}{*}{Up-Down}
&a man holding a donut \textcolor{red}{in his mouth} &a close up of \textcolor{red}{a giraffe with its wings}&\textcolor{red}{a group of people} standing in a room &a giraffe standing \textcolor{red}{in the water of} \\
&                                                   &                                                       &  with a video game                       &\textcolor{red}{a body of water} \\
\hline
\multirow{2}{*}{CNN-GAN}
&a man with glasses is holding a  doughnut   &a close up of a giraffe with \textcolor{blue}{a bird}     & \textcolor{blue}{various pictures} of a man playing & a giraffe bending down to \textcolor{blue}{drink water} \\
&                                           &\textcolor{blue}{in the background}                          &a video game                     & \textcolor{blue}{from a body of water}\\
\hline
\multirow{2}{*}{RNN-GAN}
&a man with glasses is holding a doughnut &a close up of a giraffe \textcolor{blue}{looking}   &\textcolor{blue}{a series of pictures} of a man       &a giraffe bending down to \textcolor{blue}{drink water} \\
& \textcolor{blue}{in his hands}          & \textcolor{blue}{at the camera}                     & playing a game                &  \textcolor{blue}{from a watering hole}                                         \\
\hline
\multirow{2}{*}{ensemble}
&a man with glasses is holding a doughnut   &a close up of a giraffe \textcolor{blue}{standing} & \textcolor{blue}{a series of pictures} of a man&a giraffe is bending down to \textcolor{blue}{drink water} \\
&\textcolor{blue}{in his hands}             &\textcolor{blue}{on the ground}                   &  playing a video game   &\textcolor{blue}{from a water hole}                                 \\
\midrule[1pt]
&\includegraphics[width=4.1cm,height=2.8cm]{COCO_val2014_000000420958}  &   \includegraphics[width=4.1cm,height=2.8cm]{COCO_val2014_000000404568}  &   \includegraphics[width=4.1cm,height=2.8cm]{COCO_val2014_000000119640}  & \includegraphics[width=4.1cm,height=2.8cm]{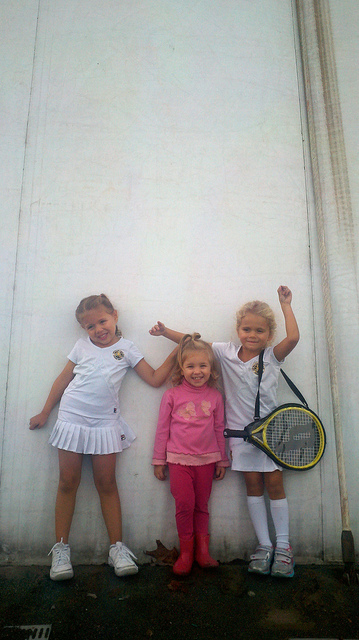}\\
\hline
\multirow{2}{*}{Up-Down}
&a view of a \textcolor{red}{mountain range in the sky} &a bird standing on \textcolor{red}{a beach in the water}&a man holding a tennis racket  &a couple of young girls holding tennis  \\
&                                                   &                                                       &  \textcolor{red}{on a tennis ball}   &rackets \textcolor{red}{on a wall } \\
\hline
\multirow{2}{*}{CNN-GAN}
&a view of an airplane wing    &a white bird standing on \textcolor{blue}{a beach}     &a man holding a tennis racket& a couple of little girls holding tennis  \\
&\textcolor{blue}{flying over the mountains}        &\textcolor{blue}{ near the ocean}  & \textcolor{blue}{on a tennis court}        &rackets\\
\hline
\multirow{2}{*}{RNN-GAN}
&a view of an airplane \textcolor{blue}{flying over}  &a bird standing on \textcolor{blue}{the beach}   &a man holding a tennis racquet       &a couple of little girls standing \textcolor{blue}{next to} \\
& \textcolor{blue}{a mountain range}          & \textcolor{blue}{ near the ocean}      & \textcolor{blue}{ on a tennis court} &  \textcolor{blue}{each other} holding tennis rackets  \\
\hline
\multirow{2}{*}{ensemble}
&a view of an airplane \textcolor{blue}{flying over}   &a white bird standing on \textcolor{blue}{a beach} &a man holding a tennis racket&a couple of little girls holding tennis \\
&\textcolor{blue}{a mountain range}             &\textcolor{blue}{ near the ocean}                     &\textcolor{blue}{ on a tennis court}    &rackets \textcolor{blue}{in front of a wall }                                 \\
\hline
\end{tabular}}
\end{center}
\end{table*}

\end{document}